%% file: root.tex
\documentclass[letterpaper, 10 pt, conference]{ieeeconf}  % Comment this line out if you need a4paper

\IEEEoverridecommandlockouts                              % This command is only needed if 
\usepackage{xcolor}
\usepackage{siunitx}  
\usepackage{graphicx}
 \graphicspath{{./}{figures/robot_demonstration}}
\graphicspath{{figures/robot_demonstration/}}
\usepackage{amsmath}
\usepackage{amssymb}
\usepackage{latexsym}
\usepackage{url}
\usepackage{cite}
\usepackage{relsize}
\usepackage{multirow}
\usepackage{afterpage}
\usepackage{ifthen}
\usepackage{graphicx}
\usepackage{algpseudocode,algorithm,algorithmicx}
\usepackage{subfigure}
\usepackage{flushend}
\usepackage{epstopdf}
\usepackage{stackengine}
\usepackage{textcomp} % new
\usepackage[font={footnotesize}]{caption}
\usepackage{gensymb}
\usepackage[bookmarks=false, linkcolor=blue, urlcolor=blue, citecolor=blue]{hyperref} 
\usepackage{booktabs}
\usepackage{makecell}
\usepackage{hhline}
\usepackage{courier}
\usepackage{lipsum}

\usepackage{xcolor} % new

\hypersetup{
    colorlinks=true,
    linkcolor=red,
    filecolor=magenta,      
    urlcolor=blue,
    pdfstartview={FitH}    }

\usepackage{xspace}

\newcommand{\etal}{\textit{et al.}\xspace}

\title{\LARGE \bf
    Text-driven Affordance Learning from Egocentric Vision
}

\author{Tomoya Yoshida$^{1, 2}$, Shuhei Kurita$^{2}$, Taichi Nishimura$^{3}$, Shinsuke Mori$^{1}$%
\thanks{$^{1}$Kyoto University, Japan}%
\thanks{$^{2}$RIKEN, Japan}%
\thanks{$^{3}$LY Corporation, Japan}
}

\usepackage{graphicx}
\usepackage{booktabs}
\usepackage{xspace}
\usepackage{orcidlink}
\usepackage{multirow}
\usepackage{amsmath,amssymb}
\usepackage{mathtools}
\usepackage{color, colortbl}

\definecolor{LightCyan}{rgb}{0.88,1,1}
\definecolor{Gray}{gray}{0.9}

\newcommand{\datasetname}{TextAFF80K\xspace}
\newcommand{\taskname}{text-driven affordance learning\xspace}

\def\Bdma#1{\mbox{\boldmath{$#1$}}}

\usepackage{CJKutf8}
\usepackage{xcolor}

\begin{document}

\maketitle
\thispagestyle{empty}
\pagestyle{empty}

%%%%%%%%%%%%%%%%%%%%%%%%%%%%%
\input{sec/0_abstract}
\input{sec/1_introduction}

\input{sec/2_related-work}
\input{sec/3_method}

\input{sec/4_experiments}
\input{sec/5_conclusion}
%%%%%%%%%%%%%%%%%%%%%%%%%%%%%

\bibliographystyle{IEEEtran}
\bibliography{IEEEabrv, reference}

\end{document}

%% file: sec/0_abstract.tex
\begin{abstract}
Visual affordance learning is a key component for robots to understand how to interact with objects.
Conventional approaches in this field rely on pre-defined objects and actions, falling short of capturing diverse interactions in real-world scenarios.
The key idea of our approach is employing textual instruction, targeting various affordances for a wide range of objects. This approach covers both hand-object and tool-object interactions.
We introduce \textit{text-driven affordance learning}, aiming to learn contact points and manipulation trajectories from an egocentric view following textual instruction. 
In our task, contact points are represented as heatmaps, and the manipulation trajectory as sequences of coordinates that incorporate both linear and rotational movements for various manipulations.
However, when we gather data for this task, manual annotations of these diverse interactions are costly.
To this end, we propose a pseudo dataset creation pipeline and build a large pseudo-training dataset: \datasetname, consisting of over 80K instances of the contact points, trajectories, images, and text tuples. 
We extend existing referring expression comprehension models for our task, and experimental results show that our approach robustly handles multiple affordances, serving as a new standard for affordance learning in real-world scenarios.
\end{abstract}

%% file: sec/1_introduction.tex
\section{Introduction}
To deploy collaborative robots in household and office environments, they should understand how to handle objects to perform human instructions effectively.
For instance, to perform \textit{``turn on the faucet''} in a kitchen, robots should understand that this action can be accomplished by rotating the faucet's handle.
\textit{Affordance}, originally proposed by Gibson \cite{affordance}, 
is a key concept for understanding how to interact with objects.
In computer vision (CV) and robotics, an affordance is often represented as contact points and manipulation trajectories \cite{hotspots, traj-aff-tpam} (Figure \ref{fig:overview} (a)).
How to acquire an affordance remains an open problem, and various approaches have been widely studied \cite{multi-label-aff, toolpart, exo, open-aff, locate, aff-demo}.

Although previous studies demonstrated strong performance and applicability to robots~\cite{vrb, affordance-cue}, they have two limitations.
First, traditional methods mainly focused on learning affordances with pre-defined objects and actions. This limits robots’ applicability in real-world scenarios because objects and actions in user instructions are diverse and it is infeasible to pre-define them.
Second, while previous studies have focused on hand-object interaction~\cite{locate, pad, large-scale-aff, vrb}, tool-object interaction is less developed~\cite{tool-obj}. Extracting the affordances of tool-object interaction is important for robots to understand the usage of tools.

\input{fig/overview}
To address these issues, in this paper, we introduce \textit{\taskname}, aiming to learn contact points and manipulation trajectory from an egocentric view following textual instruction. 
Figure \ref{fig:overview} presents the overview of our task compared with VRB~\cite{vrb}.
Our task (1) targets various affordances for diverse objects, (2) tackles both hand-object and tool-object interaction, and (3) considers both linear and rotational movements for various manipulations.
A clue for these targets is to construct a large-scale dataset for this task.

To avoid manual annotations that are costly and time-consuming, we propose an automated approach that leverages homography and off-the-shelf tools, a hand-object detector and a dense points tracker, to construct a large-scale dataset from egocentric videos.
Given an egocentric video, we first extract contact points within objects and track their movements for frames. Then we project the extracted contact points and their trajectory from frames during the interaction into frames before the interaction.
We apply this method to large-scale egocentric video datasets, Ego4D~\cite{ego4d} and Epic-Kitchens~\cite{epic}, and constructed \datasetname, which consists of over 80K tuples of images, action descriptions, contact points, and trajectories.

We extend existing referring expression models, CLIPSeg~\cite{clipseg} and MDETR~\cite{mdetr} to predict both contact points and trajectories, and train these models on the constructed datasets.
In our experiments, we evaluate the models on manually annotated datasets.
Our experimental results demonstrate two insights.
Firstly, models trained on our dataset robustly handle multiple affordances and show superior performance, particularly in tool-object interaction.
Secondly, considering both linear and rotational movements in trajectory contributes to represent complex manipulation trajectories.

%% file: fig/overview.tex
% provide photos of "open ~" and "carry ~"
\begin{figure}[t]
  \centering
  \includegraphics[width=\linewidth]{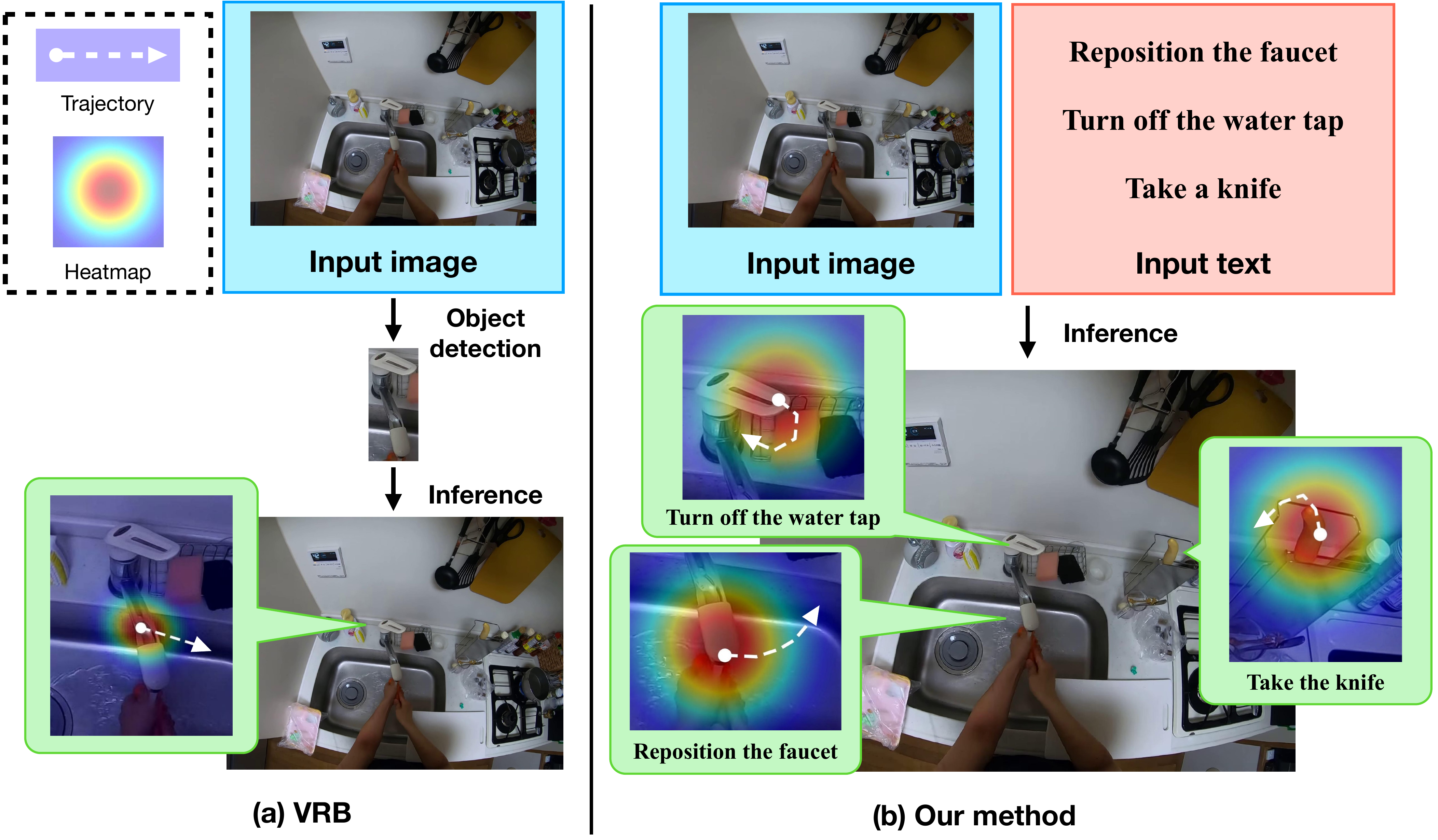}
  \vskip 0.15in
  \caption{(a) VRB~\cite{vrb}: the most closely related work. (b) Our task: Text-driven affordance learning. In our task, given an image and text, the model aims to predict the contact points and the manipulation trajectory for executing the textual instructions.}
  \label{fig:overview}
\end{figure}

%% file: sec/2_related-work.tex
\section{Related Work}

\subsection{Visual Affordance Learning}
Visual affordance learning is a key component for robots to understand how to interact with objects.
Previous studies represent visual affordance in different ways, including contact points~\cite{iit-aff, pad}, manipulation processes~\cite{oneshot-grasping, traj-aff-tpam}, and human poses~\cite{human2pose, grasppose}. These studies span various scales, from scene-level~\cite{indoorscene-affordance} and object-level~\cite{cotdet, rio, phrase-affdet, vcoco} to part-level~\cite{vrb, locate, hotspots}.
Several researchers reported that contact points and manipulation trajectories can apply to robots such as grasping tasks~\cite{vrb}. Therefore, extracting these elements has drawn much attention, and various approaches are widely studied~\cite{ego-topo, multi-label-aff,partafford,open-aff,3d-affnet}.

Early studies \cite{weakly-seg, iit-aff} focus on representing contact points as segmentation maps. Although they reported that the models are learnable to acquire fine-grained affordance of objects, pixel-level annotation is needed to train, which is costly and time-consuming. Hence, the dataset size was limited.
To address these issues, multiple approaches have been proposed such as learning from internet videos~\cite{aff-demo, hotspots}, exocentric images~\cite{locate, exo}, and building pseudo data~\cite{vrb}.
Nagarajan \etal~\cite{aff-demo} proposed a method to learn visual affordance from internet videos in weakly-supervised settings, where contact points are represented as heatmaps. 
Bahl \etal \cite{vrb} proposed VRB, which trains affordance models based on the datasets constructed automatically from egocentric videos. However, VRB has two limitations: (1) missing context of interaction indicated by contact points and (2) it focuses only on hand-object interactions and does not treat tool-object interactions.
Our work is the first attempt to resolve them and emerges as the largest in the visual affordance learning domain and contains diverse objects and actions as shown in Table \ref{tab:stats}.

\subsection{Egocentric Vision}

Egocentric vision (first-person vision) is vital to learning activities from human perspectives. Several researchers~\cite{only-ego} reported that learned representations from egocentric vision are more effective for human-action understanding than exocentric vision.
Recently, large-scale egocentric video datasets have been developed such as Ego4D~\cite{ego4d} and Epic-Kitchens~\cite{epic}. Ego4D is a massive-scale egocentric video dataset that contains 3,679 hours of daily life activity video spanning hundreds of scenarios. Epic-Kitchens is a collection of 100 hours of daily activity in kitchen video.
We leverage these datasets and construct a dataset automatically to train pseudo labels for visual affordance learning. 
\input{tab/dataset_stats}

\subsection{Referring Expression Comprehension}

Referring expression comprehension (REC)~\cite{referitgame,flickr30k,refcoco,refcocog,Liu_2019_CVPR,phrasecut} is a task of localizing objects in images from natural language descriptions.
Although REC and this study aim to localize objects in images from text inputs, we focus on learning visual affordances that consist of two-stage predictions: contact points and trajectories.
To achieve this, we extend two state-of-the-art REC models, CLIPSeg~\cite{clipseg} and MDETR~\cite{mdetr}, to predict both contact points and trajectories.

%% file: tab/dataset_stats.tex
\begin{table}[t]
\centering
\caption{Statistics of dataset on previous studies in visual affordance learning tasks and ours. HOI and TOI denote hand-object interaction and tool-object interaction, respectively.}
\vskip 0.15in
\footnotesize
\scalebox{0.95}{
\begin{tabular}{lcccccc}
\toprule
& HOI & TOI & \#Objects & \#Actions & \#Instances\\
\midrule
PAD~\cite{pad} & \checkmark & & 72 & 31 & 4,002 \\
IIT-AFF~\cite{iit-aff} & \checkmark & & 10 & 9 & 8,835 \\
ADE-Aff~\cite{learning-act} & \checkmark & & 150 & 7  & 10,000\\
ADG20K~\cite{exo} & \checkmark & & 50 & 36 & 23,816 \\
UMD~\cite{toolpart} & \checkmark & & 17 & 7 & 30,000 \\
RGBD~\cite{large-scale-aff}& \checkmark & & 7 & 15 & 47,210\\
VRB$^\dagger$~\cite{vrb} & - & - & - & - & 54K \\
\datasetname$^\dagger$ (Ours) & \checkmark & \checkmark & \textbf{5,288} & \textbf{426} & \textbf{84,770} \\
%もしseenやunseenで消えなければ111,624件
\bottomrule
\end{tabular}
}
\label{tab:stats}
\end{table}

%% file: sec/3_method.tex
\input{fig/method}
\section{Text-driven Affordance Learning}
We introduce \taskname, aiming to learn contact points and manipulation trajectories from an egocentric view following textual instruction.
Following previous studies~\cite{vrb}, the contact points and manipulation processes are represented as heatmaps and trajectories, respectively.
Because manual annotations of them are costly, we propose an automated approach that consists of three components: interaction classification, projection of contact points, and projection of trajectories as shown in Fig \ref{fig:method}. 
Given an egocentric video, we first judge whether the target interaction is hand-object or tool-object interaction. Then we extract contact points and their trajectories from frames during the interaction. Finally, we project them into frames before the interaction.
We apply this method to large-scale egocentric video datasets, Ego4D and Epic-Kitchens, and construct \datasetname.

%\subsection{Task Formulation}
%In our task, the model predicts the interaction point and the simple trajectory of interaction from a text phrase and an egocentric view.
%Image is represented as $I \in \mathbb{R}^{c \times w \times h}$, where $c$ denotes its channels, $w$ denotes its width, and $h$ denotes its height. The contact points is represented $R \in \mathbb{R}^{w \times h}$ that are often visualized in heatmap. The simple trajectory of the interaction is modeled in a combination of rotation and translation movement as
%{
%\footnotesize
%\[
%    \centering
%    \textbf{f}(t) = (\textbf{x}_0 - \textbf{x}_0^\prime) + \textbf{R}(\theta t) \textbf{x}_0^\prime + \textbf{T}t
%\]
%}
%where $t$, $R(\theta)$, $A$, $x_0$, denotes time, rotation matrix, translation matrix, initial coordinates and axis of rotation.
%$\textbf{f}(t)$ is parametarized with five parameters: $\{\theta, a, \psi, b, \phi \}$ as
%{
%\footnotesize
%\[
%    \textbf{f}(t) =
%    \textbf{x}_0 + a
%    (\begin{bmatrix}
%     \cos(\theta t) & -\sin(\theta t) \\
%     \sin(\theta t) & \cos(\theta t)     
%    \end{bmatrix}
%    - \textbf{E})
%    \begin{bmatrix}
%        \cos(\psi) \\
%        \sin(\psi) 
%    \end{bmatrix}
%    + b
%    \begin{bmatrix}
%       \cos(\phi) \\
%       \sin(\phi)
%    \end{bmatrix}
%    t.
%\label{eq}
%\]
%}

\subsection{Pseudo-Label Creation}
\label{sec:method:dataset-creation}
Our goal is to automatically create tuples $(\Bdma{x},\Bdma{d},\Bdma{p}, \Bdma{t})$ that consist of images $\Bdma{x}$, action descriptions $\Bdma{d}$, contact points $\Bdma{p}$, and manipulation trajectories $\Bdma{t}$ from the egocentric video dataset. 
Ego4D and Epic-Kitchens contain videos with action descriptions that describe what the subject in the video does at the timestep, and two timestamps that represent when the actions start and end.
We define the timestamp of the start as $t_{obs}$ and the one of the end as $t_{inter}$, also the frame at $t_{obs}$ as $F_{obs}$ and the one at $t_{inter}$ as $F_{inter}$.

\textbf{Interaction Classification.}
\label{method:dataset:classification}
Human-object interactions are roughly classified into two types, ``hand-object interaction'' and ``tool-object interaction''. Hand-object interaction is the interaction between human hands and objects, and tool-object interaction is the interaction between hand-held tools and objects. 
We aim to judge whether the target interaction is hand-object or tool-object interaction. Additionally, in the case of tool-object interactions, we aim to identify the tool used.
Because the description may not contain tools explicitly, we concatenate it with the two preceding action descriptions and input the sequence into a large language model, LLaMA2-70B~\cite{llama2}, and judge interaction types and obtain the tool name.
To obtain precise judgment, we provide a system prompt and conduct few-shot learning.

\textbf{Projection of Contact Points.}
\label{method:dataset:project_region}
Human motion between two consecutive frames is minimal, and thus it can be represented by homography~\cite{hoi}. Consequently, it is possible to get a projection matrix by multiplying homography matrices in each consecutive frame. To get a projection matrix for video clips, we first apply an off-the-shelf hand-object detector~\cite{hod} to all of the frames in the clip, which gives us 2D object bounding boxes and both hands bounding boxes. These bounding boxes are useful as masks for obtaining precise projection matrices since moving objects in frames could be noises. 

We find correspondences using SURF~\cite{surf} algorithm and calculate the homography matrix by sampling 4 pairs of points and applying RANSAC~\cite{ransac} to maximize the number of inliers. To get contact points in $F_{inter}$, we use open-vocabulary object segmentation model\footnote{\url{https://github.com/luca-medeiros/lang-segment-anything}}. The model inputs the image $F_{inter}$ and a textual phrase indicating either ``hand'' or a specific tool name that is obtained in the previous section, then outputs the segmentation maps.
We calculate the intersection regions of the mask and an object bounding box which is obtained via the hand-object detector, resulting in contact points in $F_{inter}$.
The intersection regions are projected to $F_{obs}$ by multiplying the contact points with the homography matrix, then we fit a Gaussian Mixture Model (GMM) to the regions.
To obtain contact points in $F_{obs}$, we sample several points from the fitted GMM and employ Gaussian blur to them. 

\textbf{Projection of Trajectory.}
We use an off-the-shelf dense point tracker, CoTracker~\cite{co-tracker} to the frames from $t_{inter}$ to $t_{inter} + 1.0$, and obtain segmentation maps for each frame.
To get trajectory in $F_{obs}$, we multiply homography matrices that are computed by the same method as the previous section, resulting in trajectory $P = \{(x_0, y_0), \ldots, (x_n, y_n)\}$, where $x_i$ and $y_i$ denotes the coordinates in $F_{obs}$.
Obtained trajectories include some noise, such as shaking cameras and hands, making them unsuitable for use as supervision. Therefore, we convert them to a simple trajectory of the interaction, which is modeled in a combination of rotation and translation movement as:

{
\footnotesize
\[
    \centering
    \textbf{f}(t) = (\textbf{x}_0 - \textbf{x}_0^\prime) + \textbf{R}(\theta t) \textbf{x}_0^\prime + \textbf{T}t,
\]
}

\noindent where $t$, $R(\theta)$, $A$,  $x_0$, denotes time, rotation matrix, translation matrix, initial coordinates and axis of rotation.
$\textbf{f}(t)$ is parametarized with five parameters: $\{\theta, a, \psi, b, \phi \}$ as

{
\footnotesize
\[
    \textbf{f}(t) =
    \textbf{x}_0 + a
    (\begin{bmatrix}
     \cos(\theta t) & -\sin(\theta t) \\
     \sin(\theta t) & \cos(\theta t)     
    \end{bmatrix}
    - \textbf{E})
    \begin{bmatrix}
        \cos(\psi) \\
        \sin(\psi) 
    \end{bmatrix}
    + b
    \begin{bmatrix}
       \cos(\phi) \\
       \sin(\phi)
    \end{bmatrix}
    t.
\label{eq}
\]
}

\noindent To convert the trajectory $P$ into a time-series function $\textbf{f}(t)$, we employ curve regression that aims to minimize the $l$-2 distances between the $P$ and $\textbf{f}(t)$. 
%Table \ref{fig:training-sample} depicts the sample of collected training data, tuples of heatmaps, trajectories, images, and text.

\subsection{Network Architecture}
Input text $\Bdma{d}$ and image $\Bdma{x} \in \mathbb{R} ^ {c \times w \times h}$, where $c$ denotes its channels, $w$ denotes its width, and $h$ denotes its height, models generate contact points $P \in \mathbb{R} ^ {w \times h}$ and estimate the parameter of a trajectory function $f(t)$.
Parameters related to degrees are $[0, 2\pi]$. However, this representation intuitively has a problem with continuity. Thus, we represent the parameters by $[\cos(\theta), \sin(\theta)]^T$, resulting in the representation would be continuous.
Parameters are normalized into $[0, 1]$ under training. 
We extend existing referring expression segmentation models, MDETR~\cite{mdetr} and  CLIPSeg~\cite{clipseg}.

\textbf{MDETR.}
MDETR is an end-to-end text-modulated detector based on the DETR~\cite{detr}, and it achieved high performance on the REC and RES benchmarks such as RefCOCO/+/g~\cite{refcoco, refcocog} and PhraseCut~\cite{phrasecut}.
We extend MDETR for our task, adding a deconvolution layer for predicting contact points. Additionally, we integrate a transformer encoder with a multi-layer perceptron (MLP) for trajectory estimation.

For loss functions, we use binary cross-entropy loss for generating contact points, and mean squared error for trajectory estimation. 

\textbf{CLIPSeg.}
CLIPSeg is a referring expression segmentation (RES) model that is capable of segmenting based on an arbitrary text query involving affordances and properties. It uses the pre-trained CLIP~\cite{clip} model as its backbone, FiLM~\cite{film} conditioning to inform the decoder about the segmentation target, a linear projection, and a deconvolution. 
We extend CLIPSeg for our task by adapting an MLP to the transformer decoder's output of the CLS token. 
We utilize the same loss function in the MDETR setting.

%% file: fig/method.tex
% provide photos of "open ~" and "carry ~"
\begin{figure*}[t]
  \centering
  \includegraphics[width=\textwidth]{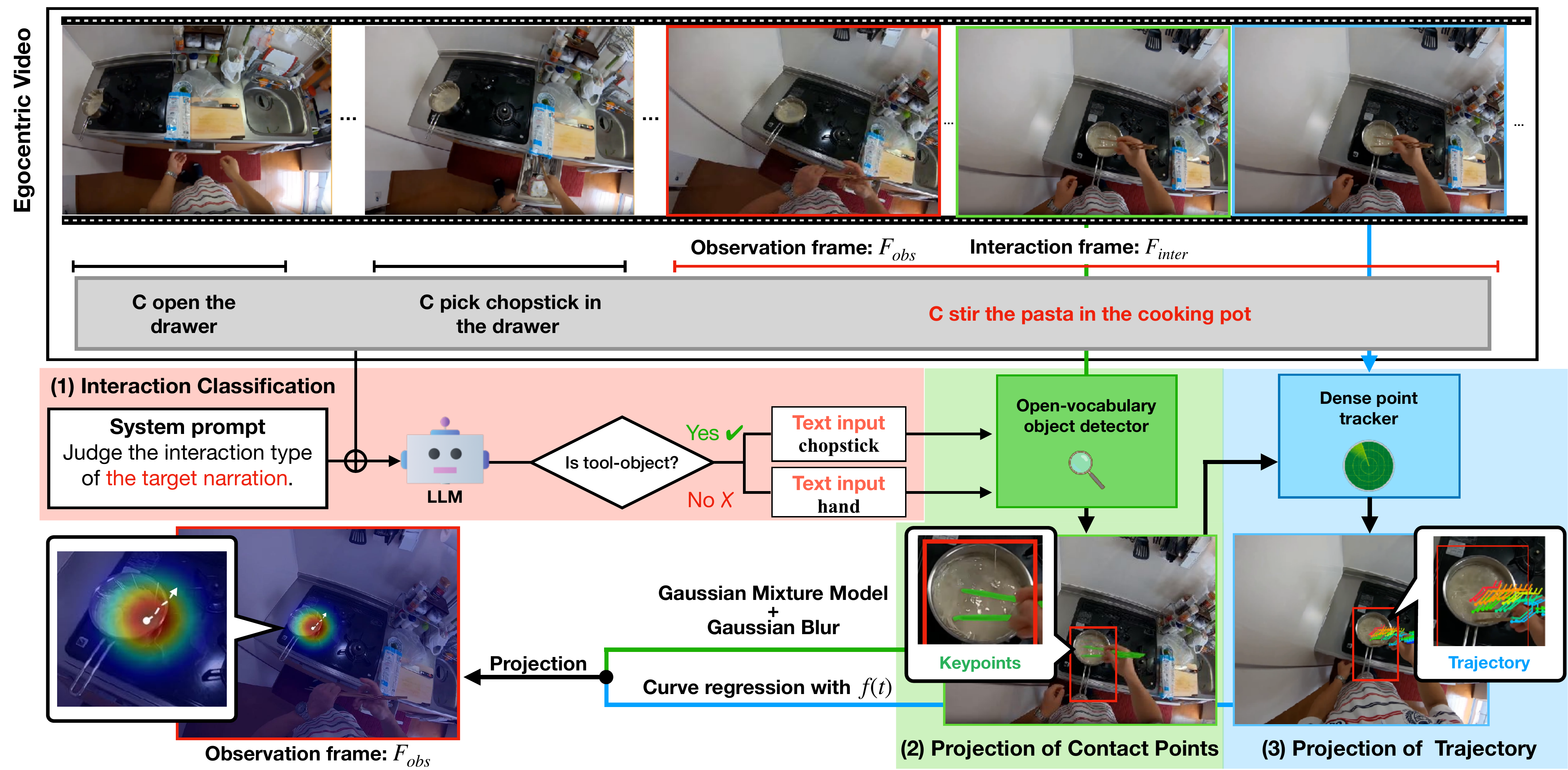}
  \vskip 0.15in
  \caption{The flow of our pseudo-label creation. This approach consists of three components: (1) interaction classification, (2) projection of contact points, and (3) projection of trajectory. Given an egocentric video, we first judge interaction type. Then we extract contact points and their trajectories from frames $F_{inter}$ during the interaction. Finally, we project them into frames $F_{obs}$ before the interaction.}
  \label{fig:method}
\end{figure*}

%% file: sec/4_experiments.tex
\input{tab/results}
\section{Experiments}
\subsection{Experimental Settings}
\textbf{Evaluation Dataset.}
To evaluate the performance of models, we collected 1,575 manually annotated data and split it into 799 validation and 776 test sets.
Figure \ref{fig:webui} depicts the manual annotation interface.
We instruct annotators to plot five key points of interaction and trajectories by referring to the action description, the interaction frame $F_{inter}$ and a one-second video starting from $F_{inter}$.
Furthermore, annotators are required to classify the type of interaction and specify the name of the tool used in the cases of tool-object interactions. 
These annotated results are converted to heatmaps and parameters of trajectories using the same approach as Section \ref{sec:method:dataset-creation}.
We filter the training set to exclude videos containing frames used in the test set collection.

\textbf{Baselines.}
We prepare the baselines of the existing visual affordance learning models that predict contact points from images: Hotspots~\cite{hotspots} and VRB~\cite{vrb}. VRB also predicts trajectories that are linear movements, which align with predicting the parameter of $\phi$.
Because these models do not take natural language inputs and their input images are expected cropped images that show one object, we extract object names from action descriptions with Spacy~\cite{spacy} and input them to an open-vocabulary object detector, Grounding DINO~\cite{g-dino}, to extract target object.
We include the result of the open-vocabulary object segmentation model of Lang-SAM\footnote{\url{https://github.com/luca-medeiros/lang-segment-anything}} combining an open-vocabulary object detection of Grounding DINO~\cite{g-dino} and Segment Anything Model~\cite{sam} for reference.

\input{fig/webui}

\textbf{Evaluation Metrics.}
Following previous work~\cite{hoi, sal-metrics}, we use five evaluation metrics, and evaluate model performance separately: contact points prediction and trajectory estimation. For contact points prediction, we use the following three metrics.
\begin{itemize}
    \item \textbf{Pearson's Correlation Coefficient (CC)}: CC computes correlation between two normalized distributions.
    \item \textbf{Similarity metrics (Sim)}: Sim~\cite{sim} measures the similarity between two histograms. Sim is computed as the sum over the minimum values of two distributions for each pixel.
    \item \textbf{AUC-Judd (AUC-J)}: AUC-J~\cite{auc-j} is a variant of AUC. The AUC evaluates the ratio of ground truth captured by the predicted map under different thresholds.
\end{itemize}

For trajectory estimation, we use the following metrics.
\begin{itemize}
    \item \textbf{Average Displacement Error (ADE)}: ADE is calculated as the $l_2$ distance between two trajectories.
    \item \textbf{Dynamic Time Warping (DTW)}: DTW~\cite{dtw} is an algorithm for measuring the distance between two sequences. It searches all possible points' alignment between two trajectories for the one with minimal distance.
\end{itemize}

For equal evaluation, we normalize predicted and ground-truth trajectories. In addition, we set initial coordinates $x_0$ to zero and evaluate trajectories independently from the heatmap.

\input{tab/detailed_results_hoi}

\input{tab/ablation}

\subsection{Results}
\label{subsec:results}
Table \ref{tab:results} presents the experimental results of our approaches and conventional baselines in the test set. 

\textbf{Contact Points Prediction.}
In terms of contact points prediction, we discover two findings. First, in hand-object interactions, VRB and Lang-SAM show better performance than CLIPSeg and MDETR, especially in Sim and CC metrics. This superiority is due to the effectiveness of the object detector of Grounding DINO because both VRB and Lang-SAM use it. Second, in tool-object interactions, CLIPSeg and MDETR outperform conventional models across all metrics. The results indicate that models that use an object detector may fall short in capturing correct objects in tool-object interactions.

\textbf{Trajecory Estimation.}
In terms of trajectory estimation, we also obtain two findings.
First, in hand-object interactions, VRB outperforms CLIPSeg and MDETR trained on \datasetname.
This observation is likely due to the nature of trajectories within hand-object interactions. Figure \ref{fig:testset_stats} presents the verb frequencies in hand-object interactions in the test set. It demonstrates that \textit{``pick''} and \textit{``take''} are dominant. These actions are often linear movements; thus VRB, which predicts only linear trajectories, achieves the highest performance.
Second, in tool-object interactions, we observe that CLIPSeg and MDETR outperform VRB, demonstrating the effectiveness of incorporating rotational movements.

\input{fig/qualitative}

\input{fig/testset_stats}
\subsection{Detailed analysis in Hand-Object Interactions}
The hand-object interaction mainly consists of \textit{``take''} and \textit{``pick''} actions. Hence, the performance in Table \ref{tab:results} is assessed primarily on these two actions and the performance on other actions remains unclear. This motivates us to further investigate the performance on each action.
We evaluate the performance based on top-2 actions: \textit{``pick/take''}, \textit{``open''}, and other actions.
Note that \textit{``pick''} and \textit{``take''} are merged because they present inherently the same actions. We call the merged one pick-like actions.

Table \ref{tab:more_detailed_analysis_hoi} presents the results, indicating that although VRB and Lang-SAM achieve better than CLIPSeg and MDETR on pick-like actions, MDETR and CLIPSeg achieve superior performance in other actions.
The results suggest that the object detector of Grounding DINO performs well in pick-like actions. However, it does not work well on other actions. MDETR and CLIPSeg are successful in capturing the affordances of these actions.

The performance of trajectory estimation also shows similar tendencies. Although VRB outperforms CLIPSeg and MDETR in pick-related actions in both metrics, MDETR and CLIPSeg outperform VRB in other actions. This is in line with the discussion in Section \ref{subsec:results} that pick-related actions often involve linear movements. Incorporating rotational movement into consideration enhances the performance of trajectory estimation for other actions.

\subsection{Ablation Study}
To validate the effectiveness of textual input, we conduct an ablation study on the extended MDETR that achieved high performance in the main result demonstrated in Table \ref{tab:results}.
%In ablation study we conduct three experiments by removing some textual components, either verbs, object names, or the whole text, from the textual input to the model.
In our ablation study, we perform three experiments to assess the impact of removing specific textual components on the model's performance. These experiments involve omitting verbs, object names, or the entire text from the model's textual input.
Table~\ref{tab:ablation} presents the ablation experiment results. The performance of predicting contact points degrades in all three ablation settings. This suggests that all of the components are essential to predict contact points accurately to some extent.
However, in terms of trajectory estimation, removing object names slightly contributes to improving the performance.
This indicates that verbs in the textual input are enough to estimate valid trajectories with the MDETR model. For example, pick-like actions tend to present a similar linear trajectory regardless of objects. Our future work is to develop a model sensitive to objects to predict trajectories accurately.

\subsection{Qualitative Analysis}
Figure \ref{fig:qualitative} visualizes contact points and trajectories of VRB, MDETR, and ground truth. Note that (a) and (b) present hand-object interactions while (c) and (d) show tool-object interactions.
In hand-object interactions, both MDETR and VRB successfully localize the appropriate points, such as the handle of the knife in (a) and the cap of the keg in (b). In terms of trajectories, in (a), both VRB and MDETR perform accurately. However, in (b), MDETR successfully predicts a valid trajectory that captures rotational movement for \textit{``open''} that VRB fails to predict.
In tool-object interactions, we observe remarkable differences in both contact point prediction and trajectory estimation. Although VRB fails to localize contact points, MDETR predicts them accurately. In addition, we observe that MDETR predicts accurate trajectories as with the ground truth.

%% file: tab/results.tex
\begin{table*}[t]
\centering
\caption{Contact Points prediction and trajectory estimation results. The column of ``Desc.'' shows whether the model inputs textual input for inferences. The symbols $\uparrow$ and $\downarrow$ denote higher and lower values are preferable for the associated metrics, respectively.}
\vskip 0.15in
% \usepackage{multirow}
\small
\begin{tabular}{lccccccccccc}
\toprule
                        & & \multicolumn{5}{c}{Hand-Object Interactions} & \multicolumn{5}{c}{Tool-Object Interactions}  \\
                        \cmidrule(lr){3-7}\cmidrule(lr){8-12}
                        & & \multicolumn{3}{c}{Contact Points $\uparrow$} & \multicolumn{2}{c}{Trajectory $\downarrow$} & \multicolumn{3}{c}{Contact Points $\uparrow$} & \multicolumn{2}{c}{Trajectory $\downarrow$} \\
                        \cmidrule(lr){3-5}\cmidrule(lr){6-7}\cmidrule(lr){8-10}\cmidrule(lr){11-12}
                        & \multicolumn{1}{l}{Desc.} & \multicolumn{1}{l}{Sim} & \multicolumn{1}{l}{CC} & \multicolumn{1}{l}{AUC-J} & \multicolumn{1}{l}{ADE} & \multicolumn{1}{l}{DTW} & \multicolumn{1}{l}{Sim} & \multicolumn{1}{l}{CC}  & \multicolumn{1}{l}{AUC-J}  & \multicolumn{1}{l}{ADE}  & \multicolumn{1}{l}{DTW}   \\
\midrule

Hotspots~\cite{hotspots} & & 12.5 & 18.6 & 61.4 & - & - & 9.5 & 11.8 & 57.4 & - & - \\
                        VRB~\cite{vrb}      & & \textbf{37.3} & \textbf{41.7} & 81.4 & \textbf{19.4} & \textbf{9.7} & 25.9 & 27.4 & 71.6 & 28.0 & 12.4 \\
                        Lang-SAM & \checkmark  & 35.8 & 41.6 & 76.2 & - & - & 23.3 & 26.6 & 64.4 & - & - \\
                        \cmidrule{1-12}
                        CLIPSeg~\cite{clipseg}  & \checkmark & 20.7 & 29.9 & 84.5 & 24.1 & 10.6 & 27.1 & 39.9 & 89.9 & 24.3 & 11.9 \\
                        MDETR~\cite{mdetr} & \checkmark & 27.6 & 35.5  & \textbf{85.9} & 24.2 & 12.0 & \textbf{32.2} & \textbf{44.1} & \textbf{90.2} & \textbf{22.7} & \textbf{11.6} \\
                        %& CLIPSeg +dd. +ft. & \checkmark  & 24.9 & 27.5 & 82.5 & \textbf{33.0} & \textbf{38.3} & \textbf{89.0}  & 27.8 & 31.2 & 84.7  \\
\bottomrule
\end{tabular}
\label{tab:results}
\end{table*}
% ``+dd.'' represents the use of deep decoder. 

%% file: fig/webui.tex
\begin{figure}[t]
  \centering
  \includegraphics[width=\linewidth]{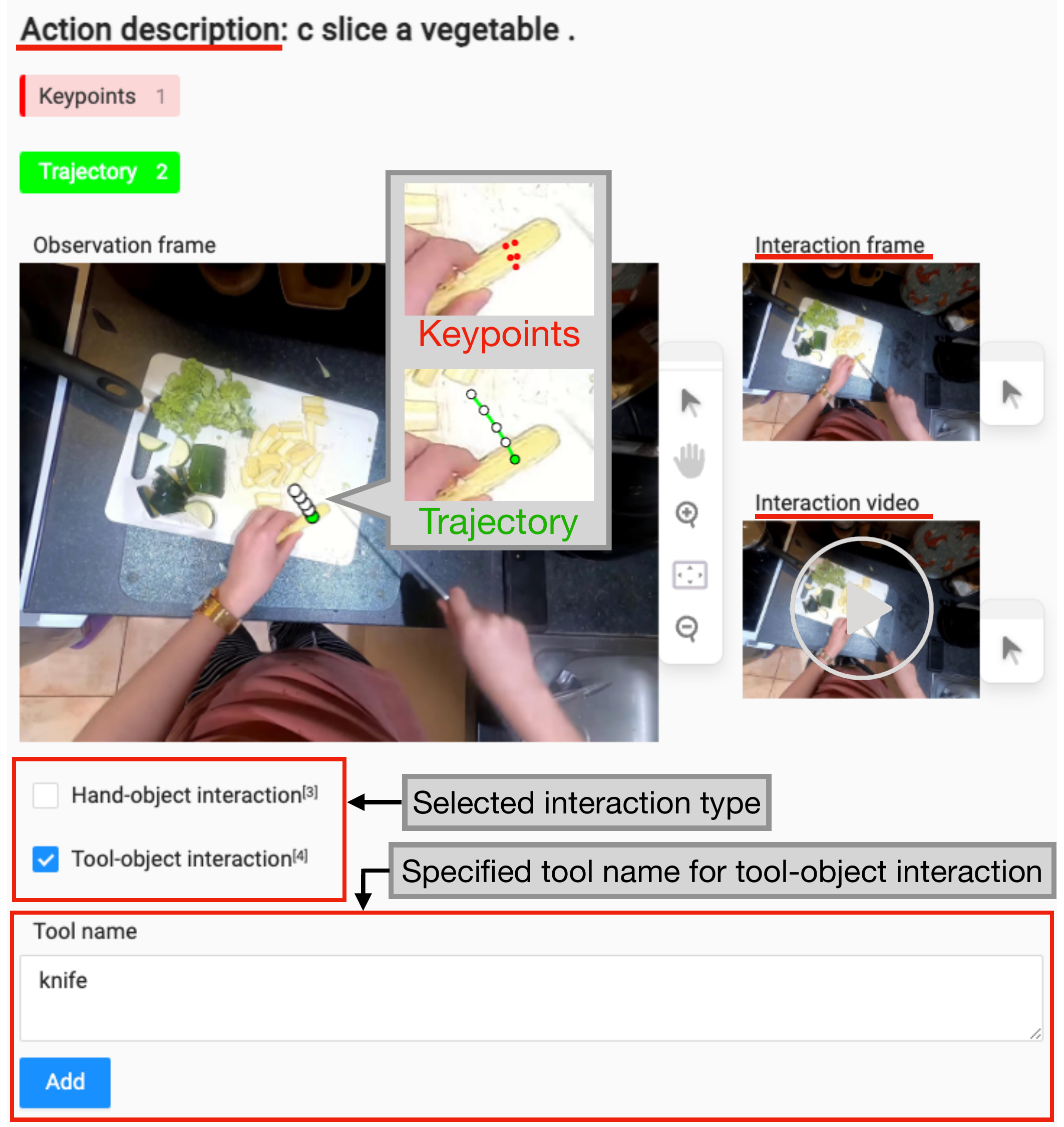}
  \vskip 0.15in
  \caption{Manual annotation tool for collecting test data.}
  \label{fig:webui}
\end{figure}

%% file: tab/detailed_results_hoi.tex
\begin{table*}[t]
\centering
\caption{Detailed analysis of contact points prediction and trajectory estimation in hand-object interactions. The column of ``Desc.'' shows whether the model inputs textual input for inferences. The symbols $\uparrow$ and $\downarrow$ denote higher and lower values are preferable for the associated metrics, respectively.
}
\vskip 0.15in
% \usepackage{multirow}
\footnotesize
\scalebox{0.95}{
\begin{tabular}{lcccccccccccccccc}
\toprule
\rowcolor{Gray}
& & \multicolumn{5}{c}{\textit{Pick and Take}} & \multicolumn{5}{c}{\textit{Open}} & \multicolumn{5}{c}{\textit{Other Actions}} \\
\cmidrule(lr){3-7}\cmidrule(lr){8-12}\cmidrule(lr){13-17} 
& & \multicolumn{3}{c}{Contact Points $\uparrow$} & \multicolumn{2}{c}{Trajectory $\downarrow$} & \multicolumn{3}{c}{Contact Points $\uparrow$} & \multicolumn{2}{c}{Trajectory $\downarrow$} & \multicolumn{3}{c}{Contact Points $\uparrow$} & \multicolumn{2}{c}{Trajectory $\downarrow$} \\
\cmidrule(lr){3-5}\cmidrule(lr){6-7}\cmidrule(lr){8-10}\cmidrule(lr){11-12}\cmidrule(lr){13-15}\cmidrule(lr){16-17}
& Desc. & Sim & CC & AUC-J & ADE & DTW & Sim & CC & AUC-J & ADE & DTW & Sim & CC & AUC-J & ADE & DTW \\
\midrule
VRB & & \textbf{42.6} & \textbf{48.3} & 82.6 & \textbf{15.2} & \textbf{8.41} & 25.8 & 27.3 & 76.2 & 25.3 & 11.4 & 31.7 & 34.6 & 83.0 & 27.4 & 12.2 \\
Lang-SAM & \checkmark & 39.6 & 45.9 & 74.7 & - & - & 23.4 & 27.2 & 71.3 & - & - & \textbf{33.3} & \textbf{39.6} & 82.0 & - & - \\
\cmidrule{1-17}
CLIPSeg & \checkmark & 19.5 & 27.3 & 83.2 & 26.4 & 13.5 & 20.5 & 33.7 & 87.9 & 22.8 & 11.9 & 23.2 & 33.0 & 86.8 & \textbf{18.9} & \textbf{10.9} \\
MDETR & \checkmark & 31.2 & 40.5 & \textbf{85.0} & 26.7 & 13.4 & \textbf{27.7} & \textbf{37.5} & \textbf{88.6} & \textbf{19.7} & \textbf{10.8} & 27.9 & 35.6 & \textbf{88.8} & 21.1 & 11.1 \\
\bottomrule
\end{tabular}
}
\label{tab:more_detailed_analysis_hoi}
\end{table*}
% ``+dd.'' represents the use of deep decoder. 

%% file: tab/ablation.tex
\begin{table}[t]
\centering
% \usepackage{multirow}
\caption{Ablation study of removing omitting verbs, objects names, or the entire text. The symbols $\uparrow$ and $\downarrow$ denote higher and lower values are preferable for the associated metrics, respectively.}
\vskip 0.15in
\small
\begin{tabular}{lccccc}
\toprule
                        & \multicolumn{3}{c}{Contact Points $\uparrow$} & \multicolumn{2}{c}{Trajectory $\downarrow$} \\
                        \cmidrule(lr){2-4}\cmidrule(lr){5-6}
                        & \multicolumn{1}{l}{Sim} & \multicolumn{1}{l}{CC} & \multicolumn{1}{l}{AUC-J} & \multicolumn{1}{l}{ADE} & \multicolumn{1}{l}{DTW} \\
\midrule                
                        MDETR & \textbf{28.7} & \textbf{37.2} & \textbf{86.8} & 22.7 & 12.0 \\
                        \; w/o verbs & 27.6 & 36.4 & 86.6 & 22.9 & 12.0 \\
                        \; w/o objects & 19.1 & 22.9 & 79.2 & \textbf{22.2} & \textbf{11.8} \\
                        \; w/o text & 15.8 & 21.4 & 79.9 & 24.5 & 12.3 \\
\bottomrule
\end{tabular}
\label{tab:ablation}
\end{table}
% ``+dd.'' represents the use of deep decoder. 

%% file: fig/qualitative.tex
% provide photos of "open ~" and "carry ~"
\begin{figure*}[t]
  \centering
  \includegraphics[width=\textwidth]{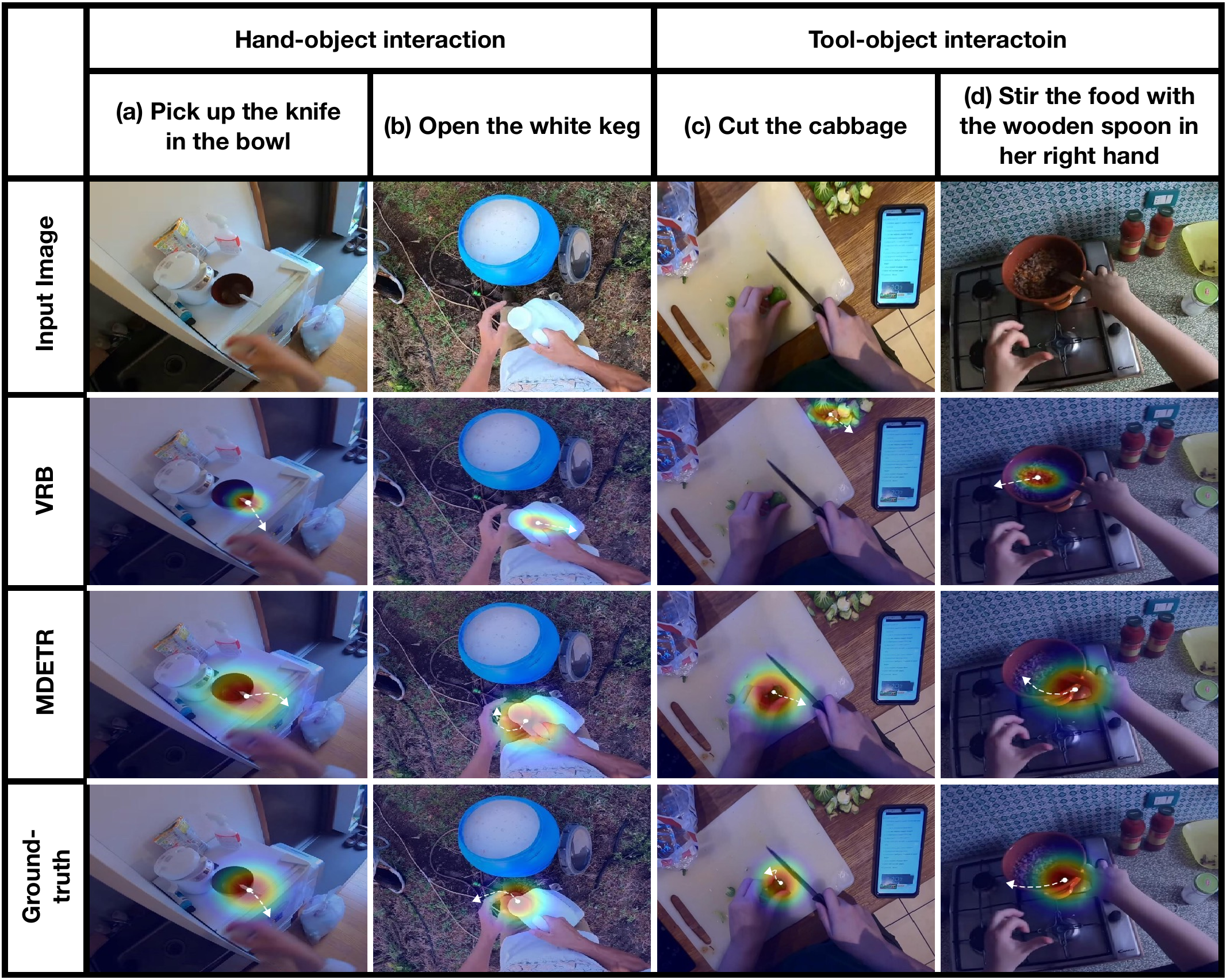}
  \vskip 0.15in
  \caption{Qualitative results. (a) and (b) depict the results of hand-object interaction, and (c) and (d) depict the results of tool-object interaction. The white dashed line presents trajectories. 
  \textbf{We cropped images for a clearer display}, but used the full images as input for the model.}
  \label{fig:qualitative}
\end{figure*}

%% file: fig/testset_stats.tex
\begin{figure}[t]
  \centering
  \includegraphics[width=\linewidth]{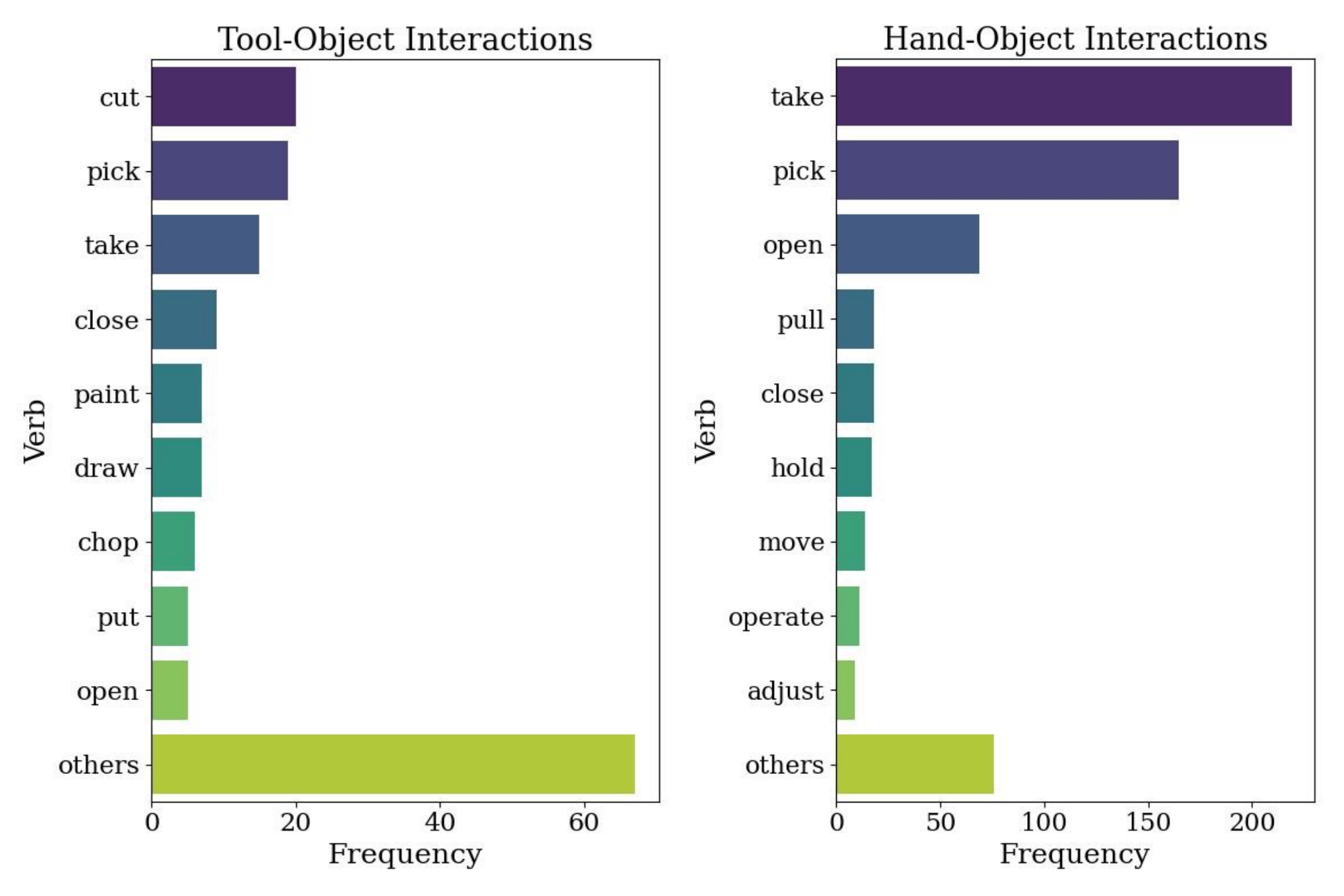}
  \vskip 0.15in
  \caption{Statistics of verb frequencies in the test set.}
  \label{fig:testset_stats}
\end{figure}

%% file: sec/5_conclusion.tex
\section{Conclusion}
In this paper, we presented \taskname, aiming to learn contact points and manipulation trajectory from an egocentric view following textual instruction.
To tackle this task, we proposed an automatic data collection pipeline, and built \datasetname that contains over 80K training instances by leveraging large-scale egocentric video datasets. 
We extended existing referring expression segmentation models and conducted experiments to verify the effectiveness of our approach.
In our experiments, models trained on our dataset robustly handle multiple affordances, and show superior performance, particularly in tool-object interactions. 
As a future direction, we plan to extend our approach to 3D environments and explore its application in robotics to address real-world challenges.